\documentclass{article}

\usepackage{arxiv}

\usepackage{url}
\usepackage{listings}
\usepackage[most]{tcolorbox}
\usepackage{cite}
\usepackage{amsmath,amssymb,amsfonts}
\usepackage{algorithmicx}
\usepackage{algorithm}
\usepackage{algpseudocode}
\usepackage{graphicx}
\usepackage{booktabs}
\usepackage{float}
\usepackage{hyperref}
\usepackage{multicol}
\usepackage{enumitem}

\usepackage{textcomp}
\usepackage{multirow}
\usepackage{xcolor}
\usepackage{adjustbox}
\usepackage[english]{babel}

\usepackage{amsmath}
\usepackage{amsfonts}
\usepackage{amssymb}

\usepackage[utf8]{inputenc}
\usepackage[T1]{fontenc}
\usepackage{tabularx}

\usepackage{listings}

\def\BibTeX{{\rm B\kern-.05em{\sc i\kern-.025em b}\kern-.08em
    T\kern-.1667em\lower.7ex\hbox{E}\kern-.125emX}}

\newtcolorbox{mybox}[2][]{%
  attach boxed title to top center
               = {yshift=-8pt},
  colback      = green!95!gray,
  colframe     = green!35!black,
  fonttitle    = \bfseries,
  colbacktitle = green!55!black,
  title        = #2,#1,
  enhanced,
}

\begin{document}

\title{\textbf{The Chronicles of RAG: The Retriever, the Chunk and the Generator}}

\author{$^*$Paulo Finardi \and $^*$Leonardo Avila \and Rodrigo Castaldoni \and Pedro Gengo  \and Celio Larcher \and Marcos Piau \and Pablo Costa \and Vinicius Caridá}

\date{\texttt{email: \{pfinardi, leonardo.bernardi.avila, castaldoniro, pedro.gengo.lourenco, celiolarcher, marcos.piau.vieira,  pablo.botton.costa, vfcarida\}@gmail.com} 
\vspace{0.4cm} \newline
\texttt{$^*$ Both authors contributed equally to this research.}}

\maketitle

\begin{abstract}

Retrieval Augmented Generation (RAG) has become one of the most popular paradigms for enabling LLMs to access external data, and also as a mechanism for grounding to mitigate against hallucinations. When implementing RAG you can face several challenges like effective integration of retrieval models, efficient representation learning, data diversity, computational efficiency optimization, evaluation, and quality of text generation. Given all these challenges, every day a new technique to improve RAG appears, making it unfeasible to experiment with all combinations for your problem. In this context, this paper presents good practices to implement, optimize, and evaluate RAG for the Brazilian Portuguese language, focusing on the establishment of a simple pipeline for inference and experiments. We explored a diverse set of methods to answer questions about the first Harry Potter book. To generate the answers we used the OpenAI's \texttt{gpt-4}, \texttt{gpt-4-1106-preview}, \texttt{gpt-3.5-turbo-1106}, and Google's \texttt{Gemini Pro}. Focusing on the quality of the retriever, our approach achieved an improvement of MRR@10 by $35.4\%$ compared to the baseline. When optimizing the input size in the application, we observed that it is possible to further enhance it by $2.4\%$. Finally, we present the complete architecture of the RAG with our recommendations. As result, we moved from a baseline of $57.88\%$ to a maximum relative score of $98.61\%$.

\end{abstract}

\section{Introduction}

The rise of Large Language Models (LLMs) has changed the way we approach Artificial Intelligence (AI) applications. Their ability to answer different user queries in different domains allow these models to show a notable performance in a wide range of tasks like translation, summarizing, question answering, and many others \cite{gpt3}. However, there are a lot of open challenges when it comes to problems that require answers based on updated information, and external data, that were not available in the training data.

In order to overcome this challenge, a technique called Retrieval Augmented Generation (RAG) \cite{rag_paper} was developed. This approach aims to solve the limitation of the need for external data, by fetching and incorporating this information in the prompt. With this, the model can generate more cohesive answers about subjects and data not seen during the training, decreasing the occurrence of hallucinations \cite{Hallucination}. Nevertheless, this approach adds a new layer of challenges since it requires the development of a trustworthy retriever pipeline, given that the quality of the final answer can be highly affected if the retrieved text is not relevant to the user query \cite{pereira2022visconde}.

The landscape of  RAG is rapidly expanding, with a constant influx of new papers introducing diverse implementations \cite{rag_survey}. Each of these variants proposes technical modifications or enhancements, such as different retrieval mechanisms, augmentation techniques, or fine-tuning methodologies. This proliferation, while a testament to the field's dynamism, presents a substantial challenge for AI practitioners. The task of methodically experimenting with, and critically evaluating, each variant's performance, scalability, and applicability becomes increasingly complex.

In this paper, we present a comprehensive series of experiments focused on the application of RAG specifically tailored for Brazilian Portuguese. Our research delves into evaluating various retrieval techniques, including both sparse and dense retrievers. Additionally, we explore two chunking strategies (naive and sentence window) to optimize the integration of retrieved information into the generation process. We also investigate the impact of the positioning of documents within the prompt, analyzing how this influences the overall quality and relevance of the generated content. Finally, our experiments extend to comparing the performance of different LLMs, notably GPT-4 and Gemini, in their ability to effectively incorporate the retrieved information and produce coherent, contextually accurate responses. This paper aims to provide valuable insights and practical guidelines for implementing RAG in Brazilian Portuguese.

Our main contributions are summarized as follows: 1) we propose a methodology to prepare a dataset in a format that allows quantifying the quality of the different steps in an RAG system. 2) We proposed a metric (maximum relative score) that allow us to direct quantify the existent gap between each approach and a perfect RAG system. 3) We discuss and compare different implementations, showing good practices and optimizations that can be used when developing a RAG system.
\section{Data Preparation}
\label{sec:dataprep}
%O conjunto de dados escolhido foi o primeiro livro da saga Harry Potter em sua versão português Brasil, esse tipo de dado é popular e o \texttt{Gemini Pro}, bem como os modelos da OpenAI conseguem responder algumas questões sobre o assunto. Além disso, utilizando o tokenizer padrão do ChatGPT \texttt{cl100k\_base} constatamos que no total existem aproximadamente $140,000$ tokens, permitindo criar prompts contendo o livro inteiro, dessa forma foi desenvolvido um conjunto de dados composto por perguntas e respostas, no qual variamos a profundidade em que as respostas estão localizadas: começo, meio ou final do prompt.

The chosen dataset was the first Harry Potter book in its Brazilian Portuguese version. This choice is motivated since it is a well known book, and both the \texttt{Gemini Pro} and OpenAI models can answer general questions on the subject. Additionally, over the application of the standard ChatGPT tokenizer \texttt{cl100k\_base}, we observed that there are approximately $140,000$ tokens in total, allowing the creation of prompts containing the entire book. Following, a dataset consisting of questions and corresponding answers was developed, with both question and answer generated by the \textit{gpt-4} model and based on a reference chunk.%, where the depth of the answer placement was varied: beginning, middle, or end of the prompt.

\begin{figure}[htb!]
\centering
\includegraphics[width=0.9\linewidth]{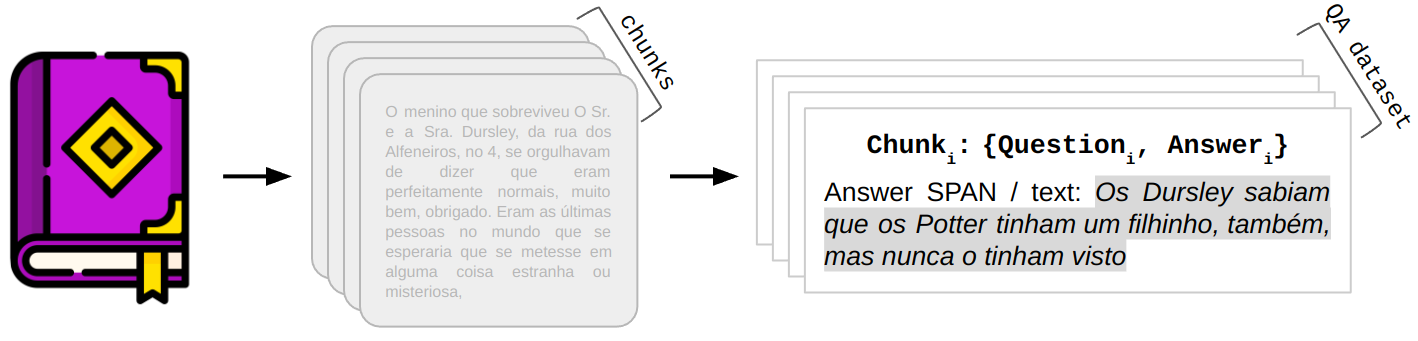} 
%\caption{De um grande documento (livro), foi criado chunks, para cada chunk foi gerado $1$ pergunta e resposta com o \texttt{gpt-4}, onde a resposta está contida no chunk.}
\caption{From a large document (book), chunks were created, and for each chunk, a question and an answer were generated using \texttt{gpt-4}, where the answer is contained within the chunk.}
\label{fig:dataprep}
\end{figure}

%A Figura \ref{fig:dataprep} detalha o processo de preparação dos dados. Inicialmente, foi escholhido, de forma arbitrária, quebrar o conjunto de dados em chunks com $1000$ tokens cada, totalizando $140$ chunks. Depois, utilizando o prompt descrito no Apêndice \ref{abstract:prompts} foi criado para cada chunk um par $\{pergunta, resposta\}$ no estilo do dataset \texttt{SQuAD} \cite{squad_dataset}, isto é, a resposta para a pergunta está presente dentro do texto de referência (chunk). 

Figure \ref{fig:dataprep} shows the data preparation process. Initially, the dataset was break into chunks with $1000$ tokens each, without overlapping, resulting in $140$ chunks. Then, using the prompt described in Appendix \ref{appendix:prompts}, a $\{question, answer\}$ pair was created for each chunk in the style of the \texttt{SQuAD} dataset \cite{squad_dataset}, meaning that the answer to the question is present within the reference text (chunk).

\section{How to Evaluate}
\label{sec:buildeval}
%Comparar dois textos nem sempre é uma tarefa trivial, por exemplo: sentence$_1 =$ \textit{"O Brasil ganhou 5 títulos mundiais de futebol."}, sentence$_2 =$ \textit{"O Brasil é pentacampeão da Copa do Mundo Fifa."}. Apesar das duas sentenças terem o mesmo sentido, metricas tradicionais como BLEU-score \cite{bleu} e ROUGE \cite{rouge} não são capazes de capturar tal semelhança, em detalhes para o exempli citado temos:

The contextual comparison of two text samples is not a straightforward task. For instance, despite the sentence$_1 =$ \textit{"Brazil has won 5 FIFA World Cup titles."}, and sentence$_2 =$ \textit{"Brazil is the five-time champion of the FIFA World Cup."} (both sentences translated into English for convenience) convey the same meaning, traditional metrics such as BLEU \cite{bleu} and ROUGE \cite{rouge} score may not be able to capture such similarity. Specifically, for the example cited:

\begin{itemize}
    %\item \texttt{BLEU score}\footnote{calculated with \url{https://huggingface.co/spaces/evaluate-metric/bleu}} $[\text{sentence}_1, \text{sentence}_2] = 0.33$
    %\item \texttt{ROUGE score} \footnote{calculated with \url{https://huggingface.co/spaces/evaluate-metric/rouge}} $[\text{sentence}_1, \text{sentence}_2] = 0.22$
    \item \texttt{BLEU score} $[\text{sentence}_1, \text{sentence}_2] = 0.33$
    \item \texttt{ROUGE score} $[\text{sentence}_1, \text{sentence}_2] = 0.22$
\end{itemize}

%A forma que tem sido mais utilizada na literatura é utilizar o \texttt{gpt-4}, com ideia similar a desenvolvida no trabalho do G-Eval \cite{g_eval}, temos um método para identificar e pontuar a comparação entre dois textos. Criamos um sistema de pontuação dividido em $4$ categorias: 

Therefore, an approach widely used in the literature is to employ \texttt{gpt-4} to provide a score based on a given prompt, a concept similar to what was done in the G-Eval work \cite{g_eval}. In this work, a scoring system divided into $5$ categories to compare two texts was devised, with scores defined as following (translated into English for convenience):

\pagebreak

\begin{itemize}
    \item score 1: The answer has no relation to the reference.
    \item score 3: The answer has little relevance but is not aligned with the reference.
    \item score 5: The answer has moderate relevance but contains inaccuracies.
    \item score 7: The answer is aligned with the reference but has minor omissions.
    \item score 10: The answer is completely accurate and aligns perfectly with the reference.
\end{itemize}

% O prompt que foi utilizado na avaliação está no Apêndice \ref{abstract:prompts}. Em nossa abordagem, utilizamos one-shot para cada categoria, no entanto, acreditaoms que a avaliação possa ficar mais robusta e determinística com adição de mais exemplos para cada categoria de pontuação, embora não tenhamos explorado essa abordagem neste momento.

The prompt used in the evaluation is shown in Appendix \ref{appendix:prompts}. Our approach uses a one-shot technique for each scoring category. Although, we believe that the evaluation could become more robust and deterministic with the addition of more few-shot examples for each scoring category, these possible variations were not explored in this work.

\subsection{Relative Maximum Score}
%Para avaliar o desempenho, avaliamos a pontuação do máximo relativo, que corresponde à pontuação máxima quando o par correto (pergunta$_i$, chunk$_i$) é fornecido ao modelo e a resposta é avaliada no sistema descrito em \ref{sec:buildeval}. Apesar de termos configurado todas as \texttt{seeds} e reutilizado os prompts da seção \ref{sec:dataprep}, o máximo relativo em nossos dados foi de aproximadamente $7.4$ pontos médios. Todos os experimentos deste trabalho foram avaliados no máximo relativo e também em termos de percentual de degradação com relação ao máximo relativo, equação \ref{eq:relative_maximum}. A Tabela \ref{tab:relative_maximum} apresenta os máximos relativos.

In order to assess performance variation for the followings experiments, we created a metric called the relative maximum score, which corresponds to the score given by a model when evaluating the correct combination of question and chunk for all pairs of a given dataset. Through this approach, it is possible to obtain the maximum score that an evaluated LLM could reach for a RAG system.

The Table \ref{tab:relative_maximum} presents results for the custom dataset created in Section \ref{sec:dataprep}, by using different LLMs to generate the answer and the \texttt{gpt-4} score system previously defined.

\begin{table}[htbp]
\caption{Relative maximum score in $140$ questions from the created QA Harry Potter dataset.}
\centering\centering\resizebox{0.42\textwidth}{!}{
\begin{tabular}{lc}
\hline
\textbf{Model}     & \textbf{Relative Maximum} \\ \hline
gpt-4              & 7.55 \\
gpt-4-1106-preview & 7.32 \\
gpt-3.5-turbo-1106 & 7.35 \\
Gemini Pro         & 7.52 \\
\hline
\end{tabular}
}
\label{tab:relative_maximum}
\end{table}

Despite configuring all the seeds and reusing the prompts, the relative maximum in our data was approximately $7.4$ average points. This points out that, even with a perfect retriever strategy, the RAG system is not able to achieve a perfect score in this dataset by using these LLM models.

For now on, all experiments in this study were assessed in terms of both the relative maximum and the percentage degradation with respect to the relative maximum, as defined in Equation \ref{eq:relative_maximum}.

\begin{equation}
\label{eq:relative_maximum}
\text{degradation score} = 1 -\frac{\text{experiment score}}{\text{relative maximum}} 
\end{equation}

With this score, we are able to address the problems regarding the retriever system itself, instead of having a vague idea of where is the main gap of our pipeline. 

\section{Introductory Experiments}
\label{sec:experiments}

% Nossos experimentos visam identificar a configuração mais eficaz para o retriever, embeddings e tipo de corte de chunk que proporcionam o melhor desempenho. Até o momento, não exploramos técnicas de prompt, embora tenhamos ciência de que a engenharia de prompt tem um impacto direto no desempenho, como demonstrado em \cite{prompt}.
%\subsection{External Knowledge}

%Nessa sessao vamos criar um baseline para as  metricas definidas em \ref{sec:buildeval}. Alem disso, vamos aplicar tecnicas de menor complexidade e comparar os resultados com o baseline. Vale comentar que não exploramos técnicas de prompt, embora tenhamos ciência de que a engenharia de prompt tem um impacto direto no desempenho, como demonstrado em \cite{prompt}. 

In this section, we will establish a baseline for the metrics defined in Section \ref{sec:buildeval}. Additionally, we will apply techniques of lower complexity and compare the results with the baseline. It's worth noting that we did not explore prompt engineering techniques, although we are aware that prompt engineering has a direct impact on performance, as demonstrated in \cite{prompt}.

\subsection{Baseline: no context}

%Sabemos que os LLMs são treinados em um conjunto de dados massivo que abrange praticamente toda a web. Essa circunstância, aliada à popularidade do universo de Harry Potter, constitui uma hipótese robusta para testarmos as perguntas de forma isolada nos modelos da OpenAI.Durante os testes com perguntas básicas, como \textit{"Quem é Harry Potter?"}, \textit{"Quem matou Dumbledore?"} e \textit{"Quais são os principais amigos de Harry Potter?"}, o ChatGPT acertou todas as respostas com precisão. No entanto, notamos que, ao lidar com perguntas mais detalhadas, o desempenho foi apenas razoável, detalhes desse experimento na Tabela \ref{tab:external_knowledge}, a seguir dois exemplos de perguntas detalhadas:

We are aware that LLMs are trained on a massive dataset that covers virtually the entire web. This circumstance, coupled with the popularity of the Harry Potter universe, forms a robust hypothesis for testing questions in isolation on OpenAI models. The answer for basic questions such as "Who is Harry Potter?", "Who killed Dumbledore?", and "What are Harry Potter's main friends?", ChatGPT were correctly answered with precision. However, we observed that when dealing with more detailed questions, the performance was only reasonable. Below are two examples of detailed questions (translated into English for convenience):

\pagebreak

\begin{itemize}
    \item \textbf{Query}: What was Harry's strategy to stop Quirrell from focusing on the mirror and what did he try to do to discover the location of the Philosopher's Stone?
    \begin{itemize}
        \item \textbf{Answer}: What occurred to Harry was to keep Quirrell talking to stop him from focusing on the mirror.
    \end{itemize}
    \item \textbf{Query}: What model of broom did Harry Potter receive and who mentioned the special circumstances to Professor Flitwick?
    \begin{itemize}
        \item \textbf{Answer}: The model of the broom that Harry Potter received is a Nimbus 2000 and it was Professor. Minerva who mentioned the special circumstances to Professor Flitwick.
    \end{itemize}
\end{itemize}

 The Table \ref{tab:external_knowledge} shows the baseline results obtained using some known LLMs for the 140 questions built as described in section \ref{sec:dataprep} and  evaluated as described in section \ref{sec:buildeval}. For this task, no retrieved context were used, only the question.

\begin{table}[htbp]
\caption{Performance of the External Knowledge experiment.}
\centering\centering\resizebox{0.5\textwidth}{!}{
\begin{tabular}{lcc}
\hline
\textbf{Model}     & \textbf{Average Score} & \textbf{Degradation} \\ \hline
gpt-4              & 5.35                   & -29.1\%\\
gpt-4-1106-preview & 5.06                   & -30.9\%\\
gpt-3.5-turbo-1106 & 4.91                   & -32.8\%\\
Gemini Pro         & 3.81                   & -50.8\%\\
\hline
\end{tabular}
}
\label{tab:external_knowledge}
\end{table}\textit{}

\subsection{Long Context}
\label{sec:long_context}
%Em comparação com os modelos GPT $1$ e $2$ \cite{gpt1, gpt2}, que lidam com até $1024$ tokens de entrada, respectivamente, o \texttt{gpt-4-1106-preview} é notável por sua capacidade de processar até $128$ mil tokens de entrada que representa um avanço exponencial na capacidade de entrada de $\approx 128$ vezes ao longo de apenas quatro anos.

In comparison to the GPT $1$ and $2$ models \cite{gpt1, gpt2}, which handle up to $1024$ input tokens, the \texttt{gpt-4-1106-preview} model stands out for its remarkable ability to process up to 128k input tokens. This represents an approximately $128$ times increase in input capacity over just four years in model development.

%Embora a arquitetura específica do \texttt{gpt-4} não tenha sido divulgada, segundo \cite{100k_context_window} é improvável que esse modelo tenha sido pré-treinado com um contexto de entrada de $128$ mil tokens. Novas técnicas pós-pré-treinamento, como ALIBI \cite{alibi} e \cite{extending_context_window}, surgiram para possibilitar a expansão da quantidade de tokens na entrada. No entanto, é importante notar que esses algoritmos podem apresentar deterioração à medida que o limite de expansão é atingido, \cite{longcontexVSretrieval}. Assim como as Redes Neurais Recorrentes (RNN) possuem um contexto teoricamente infinito, desconsiderando as limitações de desempenho e vanish-gradients, temos interesse em avaliar o desempenho do \texttt{gpt-4-1106-preview} ao longo de seus $128$ mil tokens baseada em algumas ideias do Gregory Apresentadas em \cite{greg_test_long_contex}.

The specific architecture of \texttt{gpt-4} has not been disclosed, but it is believed that this model has not been pre-trained with a 128k token input context \cite{100k_context_window}. Perhaps a post-pre-training technique could have been used, which would have made it possible to expand the number of input tokens \cite{alibi, extending_context_window}. However, it is essential to note that such a technique may show degradation as the expansion limit is reached \cite{longcontexVSretrieval}. Similar to Recurrent Neural Networks (RNNs), which theoretically have an infinite context, disregarding performance limitations and vanish-gradients, we are interested in evaluating the performance of \texttt{gpt-4-1106-preview} over its 128k tokens.

% Uma contribuição adicional é a validação do "Lost in The Middle" \cite{lostInTheMiddle}, que investiga a variação da posição da resposta ao longo do texto de referência que contém a resposta. Em nossos dados, que consistem em chunks sequenciais, conseguimos manipular a posição da resposta ao concatenar esses chunks. É importante observar que ao empilhar todos os chunks sequenciais, retomamos ao dado original. Portanto, os últimos $1, 000$ tokens do livro correspondem ao último chunk. Por exemplo, ao concatenar todos os chunks de forma sequêncial e fornecer esse grande bloco a entrada do modelo, asseguramos que a resposta estará no final de todo o input.

\begin{figure}[htb!]
\centering
\includegraphics[width=1\linewidth]{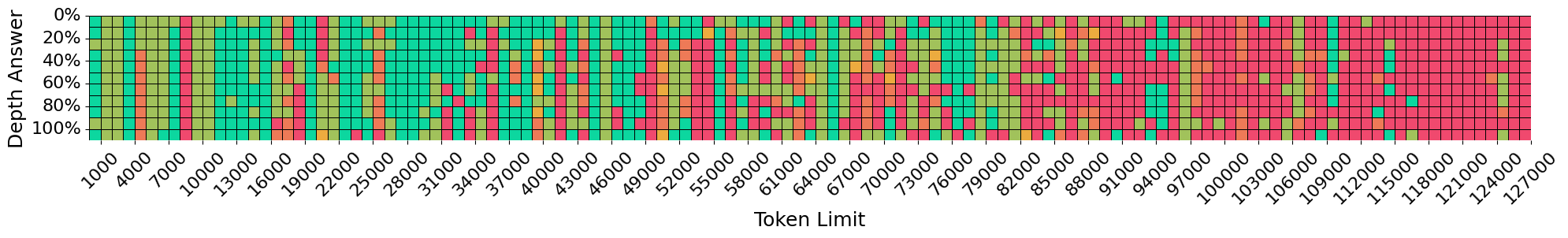} 
%\caption{Desempenho do gpt-4-1106-preview no dataset do Harry-Potter, eixo $x$: é intervalado a cada $1,000$ tokens de input do documento, eixo $y$: representa a profundidade que se encontra a resposta no documento.}
\caption{Performance of gpt-4-1106-preview on the Harry Potter dataset, x-axis: spaced at every $1,000$ tokens of input from the document, y-axis: represents the depth at which the answer is located in the document. The greener the better. Image based on Gregory repository \cite{greg_test_long_contex}.}
\label{fig:long_context}
\end{figure}

To assess the impact of the \texttt{gpt-4-1106-preview} full context capacity on the model's response, we proceed with a similar analysis of  "Lost in The Middle" \cite{lostInTheMiddle} in our dataset. This analysis explores the model output for a given question while changing the position of the answer throughout the prompt.
%Para realizar o experimento, variamos a profundidade da resposta no texto a cada intervalo de $10\%$. Assim, no eixo $y$, temos $11$ variações de profundidade de resposta, representadas por: $\{0\%, 10\%, 20\%, \dots, 90\%, 100\%\}$. No eixo $x$, representamos a quantidade de tokens utilizados como input. Por exemplo, para o ponto $(x=100,000$, $y=40\%)$, temos $(39$ chunks $+$ chunk que contém a resposta $+$ $60$ chunks) que compõem o input. A Figura \ref{fig:long_context} mostra o resultado ao longo dos $128,000$ tokens. 
To conduct this experiment, the depth of the chunk containing the answer for the question was altered in increments of $10\%$ of the total number of tokens in the context's prompt. Thus, on the y-axis, there are $11$ variations of answer depth, represented by ${0\%, 10\%, 20\%, \dots, 90\%, 100\%}$, and the x-axis represents the quantity of tokens used as input in the context, as shown in Figure \ref{fig:long_context}. The colors represents the experiment score, where the greener the better.

\pagebreak

\begin{figure}[htb!]
\centering
\includegraphics[width=0.99\linewidth]{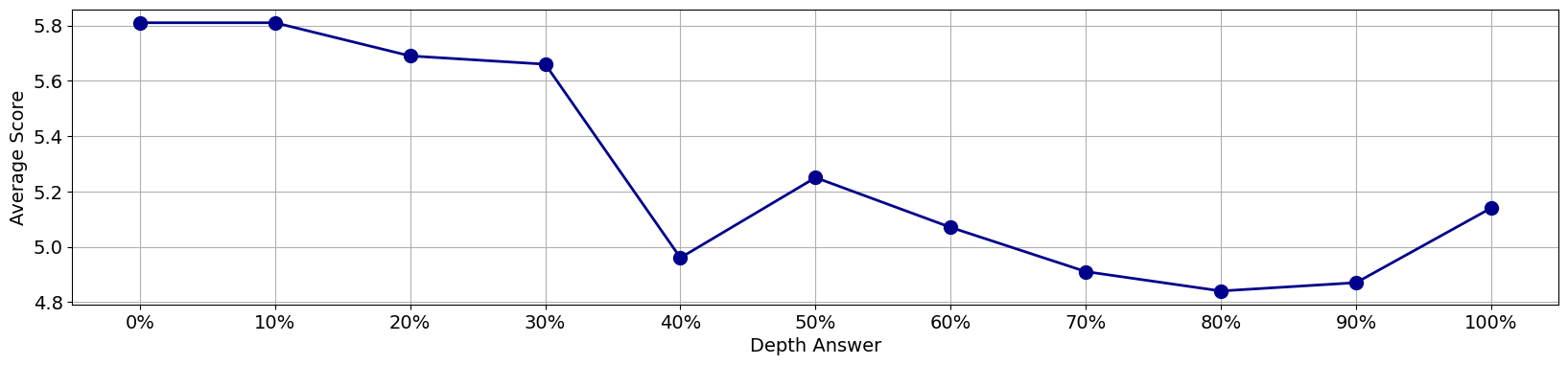} 
%\caption{Análise de desempenho da profundidade da resposta. Observamos que a as respostas que se encontram no intervalo $(40\%, $ $80\%)$ apresentam o pior desempenho, conforme constadado no artigo Lost In The Middle \cite{lostInTheMiddle}.}
\caption{Average performance analysis of \texttt{gpt-4-1006-preview} using 128k tokens context per answer depth.}
\label{fig:depth_context}
\end{figure}

For instance, for $(x=100,000$, $y=40\%)$, there are $(39$ chunks, followed by the chunk containing the answer, then by the remaining $60$ chunks, making up the $100,000$ tokens in the input context. Based on Figure \ref{fig:long_context}, we can also see when increasing the input length, we see a strong degradation in the score.
Besides that, the Figure \ref{fig:depth_context} shows that answers located in the interval of $(40\%, $ $80\%)$ exhibit the worst performance, as documented in the article "Lost In The Middle" \cite{lostInTheMiddle}.

\subsection{RAG Naive}
\label{sec:rag_naive}
% Iremos resolver o RAG de forma simples com o llama-index \cite{llamaindex}, utilizando todos os hiperparâmetros default e com o uso de recuperação de chunk por similaridade cosseno com o embedding ADA-002, a Figura \ref{fig:rag_naive} mostra o diagrama básico de como o problema é abordado, os resultados encontram-se na Tabela \ref{tab:rag_naive}.

\begin{figure}[htb!]
\centering
\includegraphics[width=0.6\linewidth]{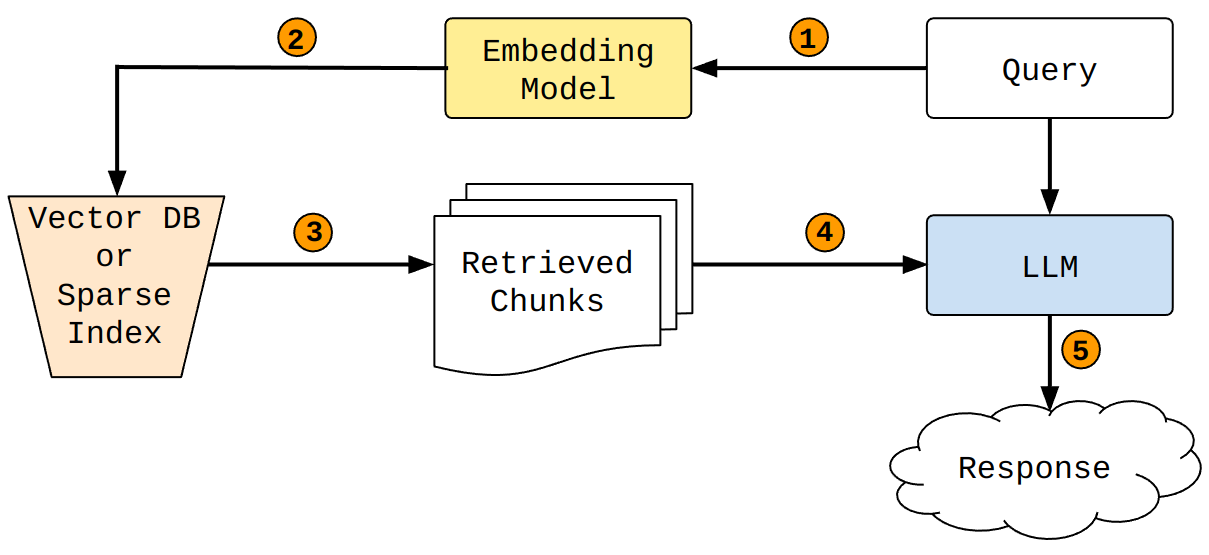} 
\caption{\texttt{1.} Pass the query to the embedding model to represent its semantics as an embedded query vector; \texttt{2.} Transfer the embedded query vector to vector database or sparse index (BM25); \texttt{3.} Fetch the top-k relevant chunks, determined by retriever algorithm; \texttt{4.} Forward the query text and the chunks retrieved to Large Language Model (LLM); \texttt{5.} Use the LLM to produce a response based on the prompt filled by the retrieved content.
 }
\label{fig:rag_naive}
\end{figure}

Initially, a straightforward approach for RAG will be done using the llama-index \cite{llamaindex}, employing all default hyperparameters and using chunk retrieval by cosine similarity with the ADA-002 embedding.  Figure \ref{fig:rag_naive} depicts the basic diagram of how the problem is addressed.

The Table \ref{tab:rag_naive} shows the average and degradation metrics for this approach using 2 retrieved chunks.

\begin{table}[htbp]
\caption{Performance of the RAG naive.}
\centering\centering\resizebox{0.5\textwidth}{!}{
\begin{tabular}{lcc}
\hline
\textbf{Model}     & \textbf{Average Score} & \textbf{Degradation} \\ \hline
gpt-4              & 6.04                   & -20\%\\
gpt-4-1106-preview & 5.74                   & -21.6\%\\
gpt-3.5-turbo-1106 & 5.80                   & -21.0\%\\
\hline
\end{tabular}
}
\label{tab:rag_naive}
\end{table}
\pagebreak

\section{Advanced Experiments}

The studies and experiments outlined in section \ref{sec:experiments} have shown unsatisfactory performance, marked by a degradation of at least $20\%$ compared to the peak relative performance. Therefore, in this section, we explore various retrieval approaches for the RAG, recognizing that the quality of the retriever is a crucial factor in enhancing performance for this type of problem. We conducted an evaluation covering both sparse and dense search, a hybrid method, and even a multi-stage architecture using a reranker.

In pursuit of code debugging flexibility and easier customization at each stage, we chose not to utilize an RAG framework (like LangChain or Llama-Index). For a comprehensive guide on debugging RAG and more details about retrieval systems, refer to \cite{build_RAG_ANYSCALE} and \cite{bert_and_beyond}.

\subsection{Retrievers}
\label{sec:retrievers}

When deploying retrieval systems, it is essential to achieve a balance between ``effectiveness'' (\textit{How good are the results returned?}) and ``efficiency'' (\textit{How much time it takes to return the results?} or \textit{How much resources are used in terms of disk/RAM/GPU?}). This balance ensures that latency, result quality, and computational budget remain within our application's required limits. This work will exclusively focus on effectiveness measures to quantify the retrievers methods quality.

In our retriever experiments, the evaluation strategy centers around assessing how well the retriever performs in retrieving relevant information based on each given query $q_i$. To achieve this, we employ the concept of recall. It is defined as the fraction of the relevant documents for a given query $q_i$ that are successfully retrieved in a ranked list $R$ \cite{bert_and_beyond}. This metric is based on binary relevance judgments, assuming that documents are either relevant or not \cite{bert_and_beyond}. In this paper, each chunk is considered a document and only the respective chunk $d_i$ is considered relevant to the query $q_i$.  While recall is easy to interpret, it does not consider the specific rank positions in which the relevant chunk appears in $R$.

To overcome this limitation, we introduce Reciprocal Rank (RR) into our analysis. In this metric, the rank of the first relevant document to the query in $R$ is used to compute the RR score \cite{bert_and_beyond}. Therefore, Reciprocal Rank offers a more nuanced evaluation by assigning a higher value when the relevant chunk is returned in the early positions of our retrievers given the respective query.

Recall and Reciprocal Rank were evaluated at a specific cutoff so the measures are presented as R@k and MRR@k. For each query, its results are evaluated and their mean serves as an aggregated measure of effectiveness of a given retriever method. The retrievers are introduced below.

In the category of sparse retrievers, we emphasize the BM25, a technique grounded in statistical weighting to assess relevance between search terms and documents. BM25 employs a scoring function that takes into account term frequency and document length, offering an efficient approach for retrieving pertinent information and is typically used as a strong baseline. However, it is exact-match based and can be powerless when query and document are relevant to each other but has no common words.

On the other hand, when exploring dense retrievers, we often encounter approaches based on the called bi-encoder design \cite{biencoder_humeau}. The bi-encoder independently encodes queries and documents, creating separate vector representations before calculating similarity. An advantage of this approach is that it can be initialized `offline': document embeddings can be precomputed, leaving only the query embedding being calculated at search time, reducing latency.

The hybrid search technique aims to leverage the best of both sparse and dense search approaches. Given a question, both searches are conducted in parallel, generating two lists of candidate documents to answer it. The challenge then lies in combining the two results in the best possible way, ensuring that the final hybrid list surpasses the individual searches. Essentially, we can conceptualize it as a voting system, where each searcher casts a vote on the relevance of a document to a given query, and in the end, the opinions are combined to produce a better result.

The multi-stage search architecture is based on the retrieve-and-rerank pipeline. In the first stage, a retriever with good recall is typically used to perform an initial filtering of the documents to be returned. From this narrowed-down list, these candidate documents are then sent to a second stage, which involves higher computational complexity, to rerank them and enhance the final effectiveness of the system.

Next, we provide more details about each retriever used.

\subsubsection{BM25}

Due to the user-friendly nature of BM25, its inclusion as a retriever method is always a welcome addition in RAG evaluations. A study that aligns with the same reasoning, albeit for a different application, can be found in \cite{bm25_baseline}, which illustrates the benefits of employing this algorithm to establish a robust baseline. Given that our data shares similarities with the \texttt{SQuAD} dataset, it is expected that many words from the query would be present in the chunk, contributing to the favorable effectiveness of BM25.

In the BM25 ranking function, \(k_1\) and \(b\) are parameters shaping term saturation and document length normalization, respectively. The BM25 formula integrates these parameters to score the relevance of a document to a query, offering flexibility in adjusting \(k_1\) and \(b\) for improving effectiveness in different retrieval scenarios.

\begin{table}[htbp]
\caption{Comparison between BM25 packages using k1=0.82 and b=0.68.}
\centering\centering\resizebox{0.6\textwidth}{!}{
\begin{tabular}{cccc}
\hline
\textbf{Recall@k}  & \textbf{rank-bm25} & \textbf{Pyserini BM25} & \textbf{Pyserini Gain (\%)} \\ \hline
3                  &  0.735              &  0.914                 & 24.3                \\
5                  &  0.814              &  0.971                 & 19.2                \\
7                  &  0.857              &  0.985                 & 14.9                \\
9                  &  0.878              &  0.985                 & 12.1                \\
\hline
\end{tabular}
}
\label{tab:bm25_compare}
\end{table}

Effectiveness is also influenced by chosen BM25 implementation. Pyserini's BM25 implementation incorporates an analyzer with preprocessing steps such as stemming and language-specific stop word removal. For the sake of comparison, we included results obtained using \texttt{rank-bm25} \cite{rankBM25}, a basic implementation without preprocessing that is widely used in Python and integrated into libraries like LangChain and Llama-index. The results can be seen in Table \ref{tab:bm25_compare}.

In this work, the Pyserini BM25 implementation \cite{pyserini} was used in all experiments considering \(k_1\)=0.82 and \(b\)=0.68.

\subsubsection{ADA-002}\label{sec:ada002}

\begin{figure}[htb!]
\centering
\includegraphics[width=0.5\linewidth]{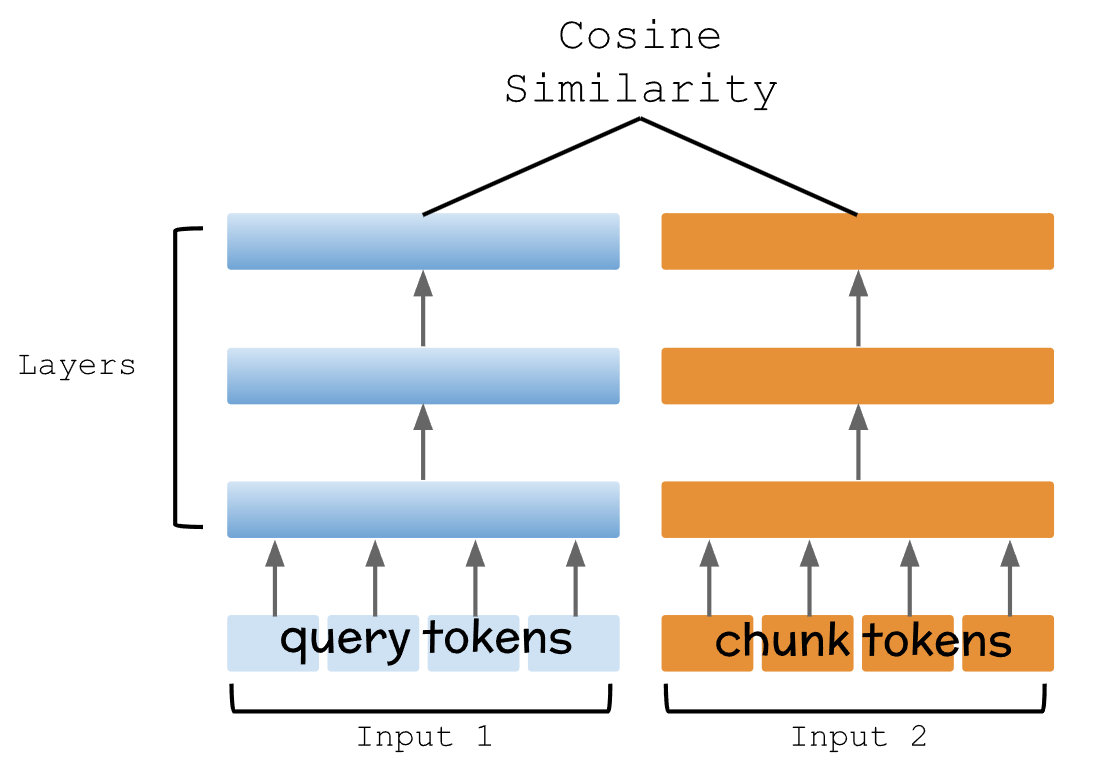} 
\caption{Bi-Encoder Architecture}
\label{fig:bi_encoders}
\end{figure}

OpenAI does not disclose extensive details about the ADA-002 architecture; however, we employed this model in retrieval as the presented bi-encoder design (Figure \ref{fig:bi_encoders}): vector representations were constructed for all available chunks, and for each input query, its embedding was computed at search time. Subsequently, the similarity between the question and chunk were assessed using cosine similarity.

Since we have no further details about ADA-002 we will refer to this approach only as dense retriever.

\subsubsection{Custom ADA-002}

The Custom ADA-002 approach was also utilized in the dense retriever configuration presented in Section~\ref{sec:ada002}. However, embedding customization played a key role in our attempt to enhance the overall representation.

Embedding customization is not limited solely to OpenAI's embeddings; it is a technique applicable to other embeddings of the same kind. There is a significant variety of approaches to optimize a matrix, with one of them being the application of Multiple Negative Ranking Loss, as presented in the article "Efficient Natural Language Response Suggestion for Smart Reply," Section 4.4 \cite{mnrloss}. However, given our current focus on simplicity, we will reserve the exploration of this technique for future work. At this moment, we choose to utilize the Mean Squared Error (MSE) Loss.

For the fine-tuning stage, it is necessary to have two types of samples:

\begin{itemize}
\item positives: [question$_i$, chunk$_i$, $label =1$]
\item negatives: [question$_i$, chunk$_j$, $label=-1$], for $i\neq j$
\end{itemize}

Often, as is the case with our dataset, only positive examples are available. However, through a simple and random shuffling, it is possible to generate negative examples. Demonstrating confidence in transfer learning, we found that a few examples were sufficient. Our final dataset consisted of approximately $400$ examples, maintaining a $1:3$ ratio between positive and negative examples.

The hyperparameters that exert the most significant impact on performance include the learning rate, batch size, and the number of dimensions in the projection matrix. The ADA-002 model has $1536$ dimensions, and the projection matrix is of size $1536\times\mathrm{N}$, where $\mathrm{N}\in {1024,2048,4096}$. In our experiments, we observed that $2048$ dimensions resulted in the best accuracy.

%\ Seria interessante colocar uma tabela ou grafico com esses resultados?

%/Este tipo de fine-tune demanda baixo recurso de GPU, com um tempo de treinamento de aproximadamente $5$ minutos usando a GPU A100. O modelo em si é simples, consistindo em uma matriz com dropout (para mitigar overfitting), seguida pela função de ativação tangente hiperbólica, que proporcionou ganhos adicionais de acurácia no conjunto de treinamento.

%/Ao analisar a similaridade cosseno entre as classes positivas e negativas, podemos observar a "sombra" compartilhada pelos histogramas. Em um cenário ideal, desejamos que as classes sejam disjuntas para garantir uma definição nítida do espaço. A Figura \ref{fig:ada_cos_sim} ilustra uma sombra significativa no embedding antes do treinamento, enquanto a Figura \ref{fig:custom_ada_cos_sim} mostra o resultado após o treinamento. Ambos os gráficos são derivados do conjunto de teste.

This type of fine-tuning requires low GPU resources, with a training time of approximately $5$ minutes using the A100 GPU. The model itself is straightforward, consisting of a matrix with dropout (to mitigate overfitting), followed by the hyperbolic tangent activation function, which provided additional accuracy gains in the training set.

% \begin{figure}[htb!]
% \centering
% \includegraphics[width=0.99\linewidth]{pics/ADA002_cosine_sim.png} 
% \caption{Similaridade cosseno das classes positivos e negativos do embedding ADA-002, observe que existe muita interseção entre as classes, test accuracy (before training): $69.5\%$}
% \label{fig:ada_cos_sim}
% \end{figure}

\begin{figure}[htb!]
\centering
\includegraphics[width=0.99\linewidth]{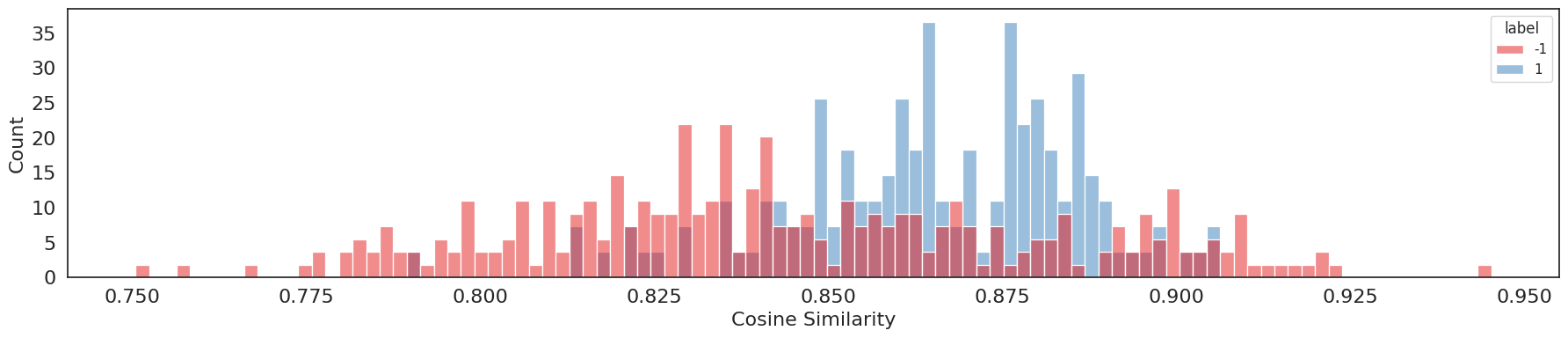} 
\caption{Cosine similarity of positive and negative classes in the ADA-002 embedding; note the significant overlap between the classes. Test accuracy \textit{(before training)}: $69.5\%$}
\label{fig:ada_cos_sim}
\end{figure}

% \begin{figure}[htb!]
% \centering
% \includegraphics[width=0.99\linewidth]{pics/ADA002_custom_cosine_sim.png} 
% \caption{Similaridade cosseno das classes positivos e negativos do embedding customizado, a interseção entre as classes é minima, , test accuracy (after training): $84.3\%$}
% \label{fig:custom_ada_cos_sim}
% \end{figure}

\begin{figure}[htb!]
\centering
\includegraphics[width=0.99\linewidth]{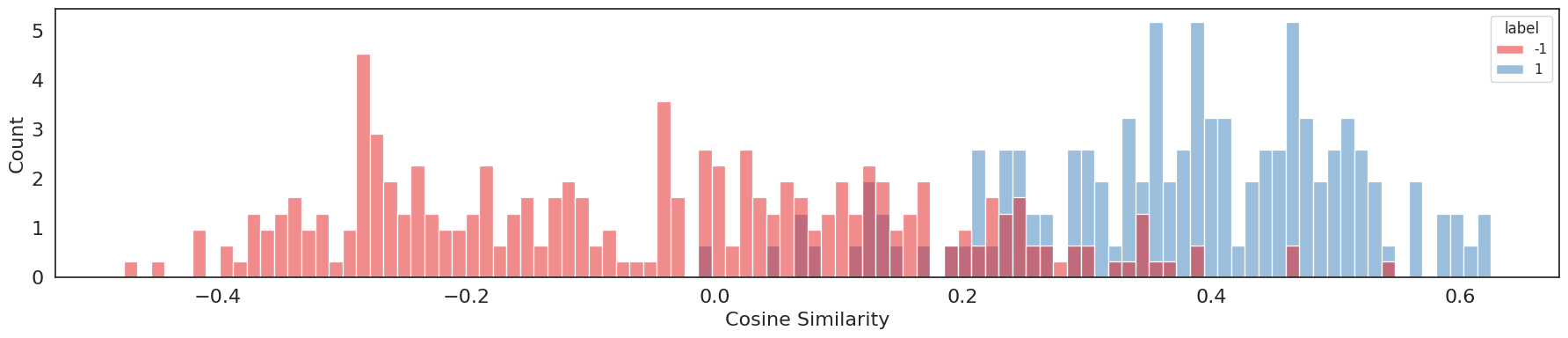} 
\caption{Cosine similarity of positive and negative classes in the customized embedding; the intersection between the classes is minimal. Test accuracy \textit{(after training)}: $84.3\%$}
\label{fig:custom_ada_cos_sim}
\end{figure}

When analyzing the cosine similarity between positive and negative classes, we can observe the "shadow" shared by the histograms. In an ideal scenario, we desire the classes to be disjoint to ensure a clear definition of space. Figure \ref{fig:ada_cos_sim} illustrates a significant shadow in the embedding before training, while Figure \ref{fig:custom_ada_cos_sim} shows the result after training. Both graphs are derived from the test set. Test accuracy also improved, leading to a better dense representation.

\subsubsection{Hybrid Search}

As stated before, hybrid search is applied when is necessary to combine results from two or more retrieval methods. A widely used algorithm to address this type of problem is known as Reciprocal Rank Fusion (RRF). For a document set $D$ and search results from different methods $r$ in $R$, for each $d$ in $D$, we can calculate the $RRF_{score} (d \in D)$ as follows \cite{hybrid_fusion}:

% \text{RRF_score}(d \in D) = \sum_{r \in R} \frac{1}{k + r(d)},
\begin{equation}
RRF_{score}(d \in D) = \sum_{r \in R} \frac{1}{k + r(d)},
\end{equation}

% Sendo que $1/r(d)$ é conhecido por reciprocal rank e $r(d)$ é a posição em que o documento $d$ foi retornado pelo mecanismo de busca $r$. O termo $k$ é adicionado para ajudar a controlar sistemas outliers.

Considering that $1/r(d)$ is known as the reciprocal rank, where $r(d)$ represents the position at which the document $d$ was retrieved by the search mechanism $r$. The term $k$ is introduced to assist in controlling outlier systems \cite{hybrid_fusion}.

\begin{figure}[htb!]
\centering
\includegraphics[width=0.99\linewidth]{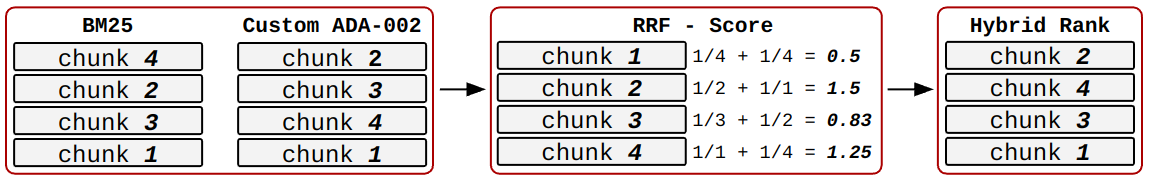} 
\caption{Hybrid Search schema with k=1.}
\label{fig:hybrid}
\end{figure}

% A Figura \ref{fig:hybrid} representa essa abordagem tomando $k=1$. No exemplo, temos 4 chunks que foram retornados em diferentes ordens por dois métodos de busca, o BM25 (busca esparsa) e o Custom ADA-002 (busca densa). Para cada chunk é calculado o reciprocal rank score. Esses valores são então somados, formando um novo score. A lista final híbrida é uma ordenação dos chunks que utiliza esse novo score de forma decrescente.

% Em nossas abordagens, apenas o BM25 do Pyserini foi testado como buscador esparso, enquanto que ADA-002 e Custom ADA-002 foram testados como buscadores densos. A combinação híbrida que apresentou melhor resultado foi a que utilizou BM25 e Custom ADA-002.

Figure \ref{fig:hybrid} shows how to calculate $RRF_{score}$ for $k=1$. In the example, we have four chunks that were retrieved in different orders by two search methods, BM25 (sparse search), and Custom ADA-002 (dense search). The reciprocal rank score is calculated for each chunk. These values are then summed, creating a new score. The final hybrid list is an ordering of chunks that uses this new score.

\begin{table}[htbp]
\caption{Retriever comparison. Where MRR is the Mean Recriprocal Rank metric and R@k is the Recall.}
\centering\centering\resizebox{0.7\textwidth}{!}{
\begin{tabular}{cccccc}
\hline
\textbf{Metric}  &  \textbf{Hybrid-BM25-ADA-002} & \textbf{Hybrid-BM25-Custom ADA-002}     \\ \hline
MRR@10             & 0.758                         & 0.850                              \\
R@3                & 0.829                         & 0.921                              \\
R@5                & 0.879                         & 0.943                              \\
R@7                & 0.921                         & 0.964                              \\
R@9                & 0.957                         & 0.979                              \\
\hline\end{tabular}}
\label{tab:topk_retrieved_ada}
\end{table}

In our experiments, only Pyserini's BM25 was tested as the sparse retriever, while ADA-002 and Custom ADA-002 were tested as dense retrievers. The hybrid combination that yielded the best results was the one that used BM25 and Custom ADA-002.

\subsubsection{Reranker}
\begin{figure}[htb!]
\centering
\includegraphics[width=0.8\linewidth]{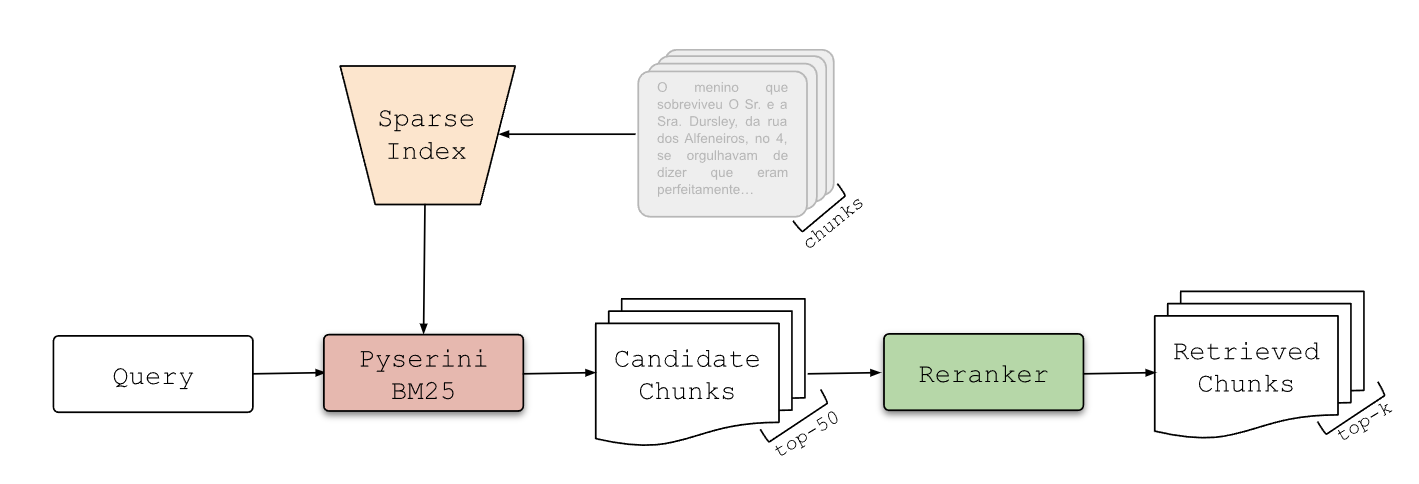} 
\caption{Reranker Pipeline}
\label{fig:reranker}
\end{figure}

The fundamental idea underlying multi-stage ranking is to divide document ranking into a sequence of stages. After an initial retrieval, which usually involves a sparse retriever or dense retriever, each subsequent stage re-evaluates and reranks the set of candidates forwarded from the preceding stage. Figure \ref{fig:reranker} represents a multi-stage pipeline where Pyserini BM25 performs the first stage and the candidate chunks are then re-evaluated by the reranker. After that, the final reranked list called retrieved chunks is presented as final result formed by $k$ chunks.

Transformers-based models are commonly employed as rerankers, leveraging their capability to enhance the effectiveness of information retrieval systems by capturing intricate relationships and contextual information within documents and queries.

\begin{figure}[htb!]
\centering
\includegraphics[width=0.5\linewidth]{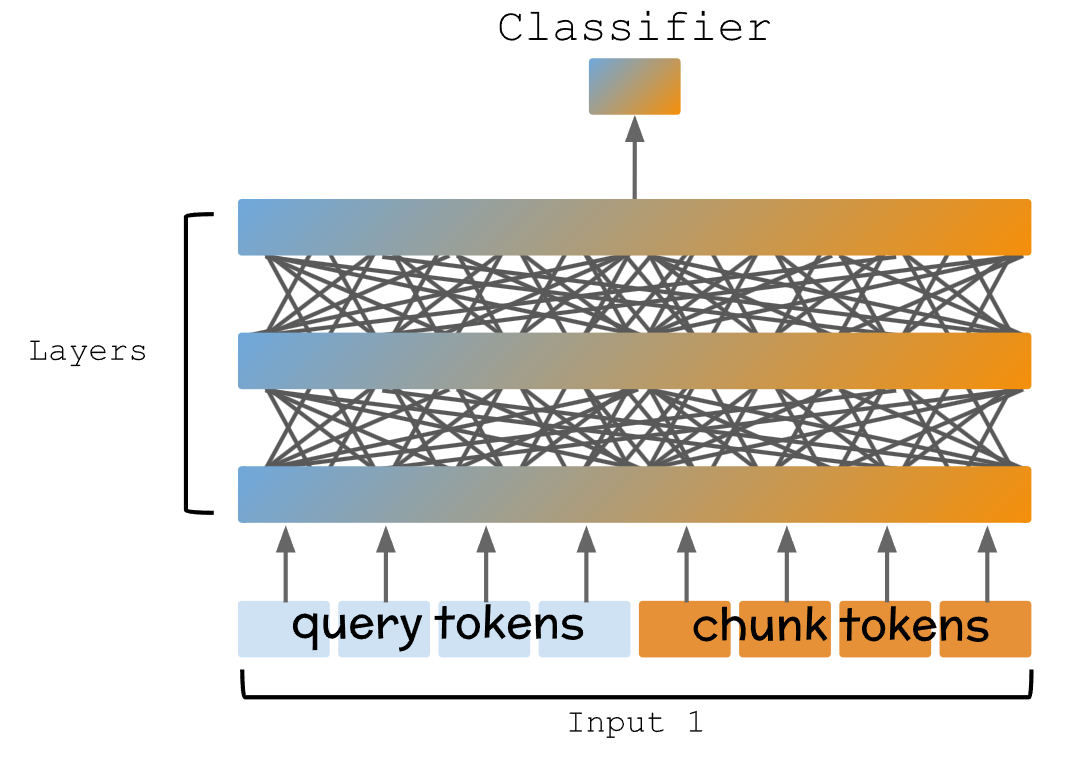} 
\caption{Cross-Encoder}
\label{fig:cross_encoder}
\end{figure}

The initial utilization of transformers within a multi-stage ranking framework was presented in the study \cite{monobert}. Their proposed model, known as monoBERT, transforms the ranking process into a relevance classification problem. It achieves this by sorting texts based on the conditional probability \( P(\text{Relevant} = 1 \,|\, d_i; q) \), where \( q \) is the query and \( d_i \) represents documents \cite{bert_and_beyond}. The model simultaneously processes both queries and documents. This simultaneous processing leads to a more enriched interaction between them, often resulting in improved effectiveness \cite{cross_rank_ref_1}, \cite{cross_rank_ref_2}. However, this neural models have a substantial number of parameters and the scoring of query-document pairs occurs at inference time. This, in turn, increases computational costs and latency. This kind of model is also known by cross-encoder. Refer to Figure \ref{fig:cross_encoder} for an illustration of the query-document pair processing.

\begin{figure}[htb!]
\centering
\includegraphics[width=0.99\linewidth]{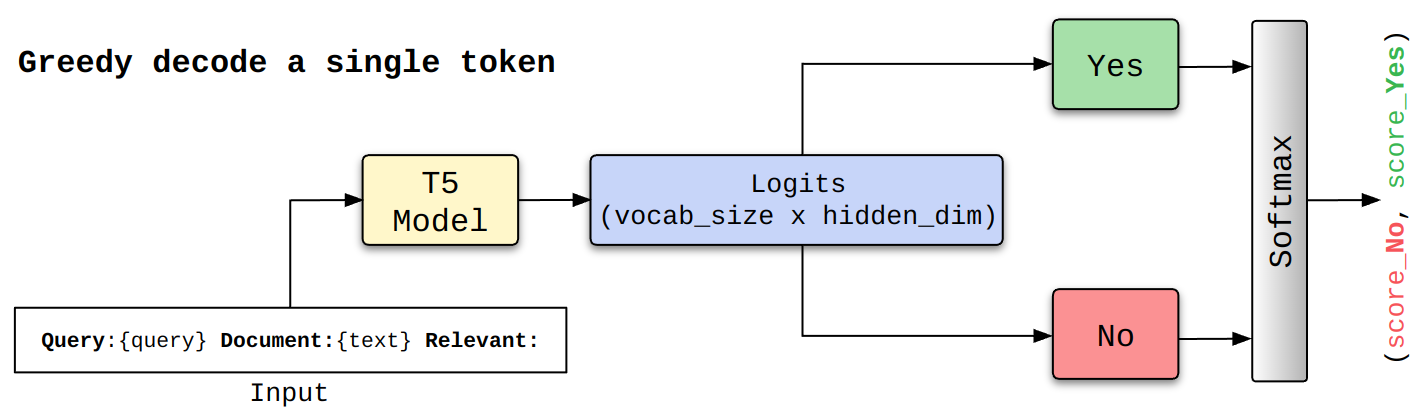} 
\caption{monoT5's inference process}
\label{fig:monot5}
\end{figure}

In the other hand, monoT5 is a sequence-to-sequence reranker \cite{monot5} that uses T5 models \cite{t5-google} to generate relevance scores between a pair of query and document. T5 models treat all tasks as \textit{text-to-text}, requiring some adaptations for the query/document similarity task. During training, the format \texttt{\textquotesingle Query: \{query\} Document: \{document\} Relevant:\textquotesingle} is used, with labels \texttt{yes} if the document is relevant to the query, and \texttt{no} otherwise. At inference time, the same format as the training data is used to format pairs of queries and documents to feed the model, and a single-token greedy decode is performed. The score is then obtained by calculating the softmax value considering only the tokens \texttt{no} and \texttt{yes}, and selecting the value corresponding to the \texttt{yes} token. Note that the tokens \texttt{no} and \texttt{yes} are the tokens used in the version provided by \cite{mmarco}, but the publication that introduced the monoT5 architecture, \cite{monot5}, uses the tokens \texttt{false} and \texttt{true}. Figure \ref{fig:monot5} contains an illustration of the monoT5 architecture's inference process.

In our experiments, Pyserini BM25 was employed as first-stage retrieval, retrieving 50 documents to be reranked in the second stage. In the second stage, we utilized the model \texttt{unicamp-dl/mt5-base-en-pt-msmarco-v2}, a sequence-to-sequence reranker trained on pairs of queries and documents in English and Portuguese from the dataset \cite{mmarco}.

\subsection{Retrievers Results}

The results achieved with the various retrievers are presented in Table \ref{tab:topk_retrieved}.
 
\begin{table}[htbp]
\caption{Retriever comparison.}
\centering\centering\resizebox{1\textwidth}{!}{
\begin{tabular}{ccccccc}
\hline
\textbf{Metric} & \textbf{ADA-002} & \textbf{Custom ADA-002} &  \textbf{Hybrid-BM25-ADA-002} &  \textbf{Hybrid-BM25-Custom ADA-002} & \textbf{BM25}  & \textbf{BM25 + Reranker} \\ \hline
MRR@10           & 0.565            & 0.665                   & 0.758                        & 0.850                               & 0.879        & 0.919               \\
R@3              & 0.628            & 0.735                   & 0.829                        & 0.921                               & 0.914        & 0.971               \\
R@5              & 0.692            & 0.835                   & 0.879                        & 0.943                               & 0.971        & 0.985               \\
R@7              & 0.750            & 0.871                   & 0.921                        & 0.964                               & 0.985        & 0.992               \\
R@9              & 0.814            & 0.921                   & 0.957                        & 0.979                               & 0.985        & 1                   \\
\hline\end{tabular}
}
\label{tab:topk_retrieved}
\end{table}

The multi-stage pipeline was able to achieve the best results in MRR@10 and Recall@k.

\section{Conclusions}

% A implementação do RAG enfrenta desafios como a integração eficaz de modelos de recuperação, aprendizado eficiente de representação, diversidade de dados, otimização da eficiência computacional, avaliação e qualidade de geração de texto. Diante desses obstáculos em constante evolução, este artigo propõe boas práticas para implementação, otimização e avaliação do RAG em um dataset em Português Brasil, enfocando um pipeline simplificado para inferência e experimentação.

The implementation of RAG systems faces challenges such as the effective integration of retrieval models, efficient representation learning, the diversity of data, the optimization of computational efficiency, evaluation, and text generation quality. Faced with these constantly evolving obstacles, this article proposes best practices for the implementation, optimization, and evaluation of RAG on a Brazilian Portuguese dataset, focusing on a simplified pipeline for inference and experimentation.

% Até aqui você foi apresentado os principais componentes e métodos, juntamente com seus resultados e lacunas. Nesta seção, discutiremos os pontos-chave que levam à melhoria de desempenho nas aplicações do RAG. Iniciaremos discutindo a relação entre a qualidade do retriever e o desempenho alcançado, no qual nossa abordagem obteve uma melhoria significativa de $35.4\%$ em relação ao baseline. Em seguida, abordaremos o impacto do tamanho da entrada na performance. Nesse domínio, observamos que é possível aprimorar em $2.4\%$ a melhor estratégia de recuperação de informações ao otimizarmos o tamanho da entrada. Por fim, apresentaremos a arquitetura completa do RAG com nossas recomendações. Ao avaliarmos a acurácia final da nossa abordagem, alcançamos $98.61\%$, o que representa uma melhoria de $40.73$ em less degradation score em relação ao baseline.

So far, we have introduced the main components and methods along with their results and gaps. In this section, we will discuss key points that contribute to the performance improvement in RAG applications. We will start by discussing the relationship between the quality of the retriever and the achieved performance, in which our approach showed a significant improvement of MRR@10 by $35.4\%$ compared to the baseline. Next, we will address the impact of input size on performance. In this domain, we observed that it is possible to enhance the best information retrieval strategy by $2.4\%$ through input size optimization. Finally, we will present the complete architecture of RAG with our recommendations. When evaluating the final accuracy of our approach, we reached $98.61\%$, representing an improvement of $40.73$ in less degradation score compared to the baseline.

\subsection{Retriever Score versus Performance}
% Como analisado na seção \ref{sec:retrievers}, o desempenho da recuperação de informações/chunks, medido pela métrica MRR@10, varia entre $(0.565, 0.919)$, conforme detalhado na Tabela \ref{tab:topk_retrieved}. Essa variação representa aproximadamente $35.4\%$. É importante destacar que o desempenho do RAG é diretamente afetado pela qualidade do retriever. A Figura \ref{fig:piau_png} ilustra os métodos de retriever estudados, classificados por score de degradação.

\begin{figure}[htb!]
\centering
\includegraphics[width=0.99\linewidth]{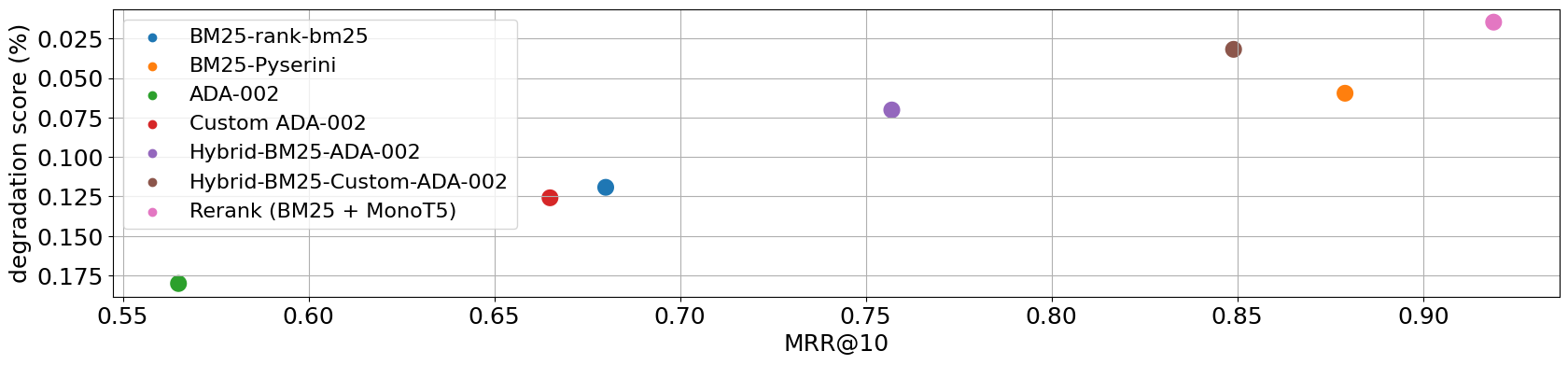} 
\caption{Retriever effectiveness vs RAG Performance. $x$ axis is the MRR@10 metric and $y$ axis is the degradation (where $0$ is the perfect scenario) score.}
\label{fig:piau_png}
\end{figure}

As mentioned in section \ref{sec:retrievers}, the performance of information/chunk retrieval, measured by the MRR@10 metric, varies between $(0.565, 0.919)$, as detailed in Table \ref{tab:topk_retrieved}. This variation represents approximately $35.4\%$. It is important to highlight that the RAG's performance is directly influenced by the quality of the retriever. Figure \ref{fig:piau_png} shows the relationship between the retrieval metric MRR@10 metric and the degradation score for the studied retrieval methods.

\subsection{Input Size versus Performance}

\begin{table}[htbp]
\caption{Performance of number of chunks retrieved with \texttt{gpt-4}.}
\centering\centering\resizebox{0.75\textwidth}{!}{
\begin{tabular}{cccccc}
\hline
\textbf{\# Retrieved Chunks}  & \textbf{ADA-002} &\textbf{Custom ADA-002} & \textbf{BM25} & \textbf{Hybrid} & \textbf{BM25 + Reranker}\\ \hline
3                             & 6.19             & 6.41                   &    7.10       & 7.31            & \textbf{7.44}   \\
5                             & 6.29             & 6.61                   &    7.32       & 7.37            & 7.43            \\
7                             & 6.42             & 6.82                   &    7.17       & 7.20            & 7.32            \\
9                             & 6.57             & 6.88                   &    7.22       & 7.34            & 7.37            \\
\hline
\end{tabular}
}
\label{tab:topk_chunks}
\end{table}

% Observamos que o melhor desempenho foi obtido com uma recuperação de $3$ chunks com a estratégia rerank, Tabela \ref{tab:topk_chunks}. O uso do cross-encoder do rerank \ref{fig:bi_and_cross_encoders} apresentou melhor recuperação da informação em nossos testes. Com essa mesma configuração o  \texttt{Gemini Pro} obteve desempenho similar ao \texttt{gpt-4}, Tabela \ref{tab:rag_opt_degradation}. 

We observed that the best performance was achieved with the retrieval of $3$ chunks using the retrieve-and-rerank strategy, as shown in Table \ref{tab:topk_chunks}. The use of a reranker (Figure \ref{fig:reranker}) showed improved information retrieval in our tests. With this same configuration, the \texttt{Gemini Pro} achieved performance similar to \texttt{gpt-4}, as indicated in Table \ref{tab:rag_opt_degradation}.

\begin{table}[htbp]
\caption{Performance of best retriever RAG.}
\centering\centering\resizebox{0.65\textwidth}{!}{
\begin{tabular}{llcc}
\hline
\textbf{Model} & \textbf{Retriever Method}     & \textbf{\# Retrieved Chunks} & \textbf{Degradation} \\ \hline
gpt-4 & ADA-002                       & 9                            & -13.0\%\\
gpt-4 & ADA-002 Custom                & 9                            & -8.8\%\\
gpt-4 & BM25                          & 5                            & -3\%\\
gpt-4 & Hybrid                        & 5                            & -2.3\%\\
gpt-4 & \textbf{BM25 + Reranker}              & \textbf{3}                   & \textbf{-1.4\%}\\
Gemini Pro & BM25 + Reranker                  & 3                           & -2.2\%\\
\hline
\end{tabular}
}
\label{tab:rag_opt_degradation}
\end{table}
% Apesar do recall perfeito para $9$ chunks, conforme evidenciado na Tabela \ref{tab:topk_retrieved} utilizar um input com $9000$ tokens, $6000$ a mais do que o melhor caso ($3$ chunks) não resultou na melhor performance. Conforme discutido na Seção \ref{sec:long_context}, a qualidade do RAG está diretamente relacionada ao tamanho do input e à posição onde a resposta se encontra. Portanto, os resultados finais confirmam as observações realizadas em nossos experimentos.

% Além disso, do ponto de vista financeiro, é crucial evitar sobrecarregar o LLM com uma grande quantidade de tokens de entrada, uma vez que o custo também é baseado na quantidade de texto da entrada. 

% Destacamos que os resultados obtidos nesse trabalho não podem ser considerado como uma generalização para outros datasets. Exploration Data Analysis e uso de boas práticas de retriever, conforme apresentado em \cite{retrieverSurvey}, é sempre um caminho sólido para bons resultados.

Despite achieving perfect recall for $9$ chunks, as evidenced in Table \ref{tab:topk_retrieved}, using an input with $9000$ tokens, $6000$ more than the best scenario ($3$ chunks), did not result in the best performance. As discussed in Section \ref{sec:long_context}, the quality of RAG is directly related to the input size and the position where the answer is located. Therefore, the final results confirm the observations made in our experiments.

Moreover, from a cost perspective, it is crucial to avoid overloading the LLM with a large number of input tokens, as the cost is also based on the amount of input text.

It is important to note that the results obtained in this study cannot be considered as a generalization for other datasets. Exploratory Data Analysis and the use of good retriever practices, as presented in \cite{retrieverSurvey}, are always a solid path to achieving good results.

\subsection{Final Results}

% Apesar de este trabalho ser fundamentado em um único conjunto de dados, é sempre fundamental ressaltar a importância da \textit{qualidade dos dados}. De maneira simplificada, como ilustrado na Figura \ref{fig:rag_core_points}, a \textit{qualidade dos dados} no RAG pode ser dividida em: Input, Retriever e Evaluation.

\begin{figure}[htb!]
\centering
\includegraphics[width=0.99\linewidth]{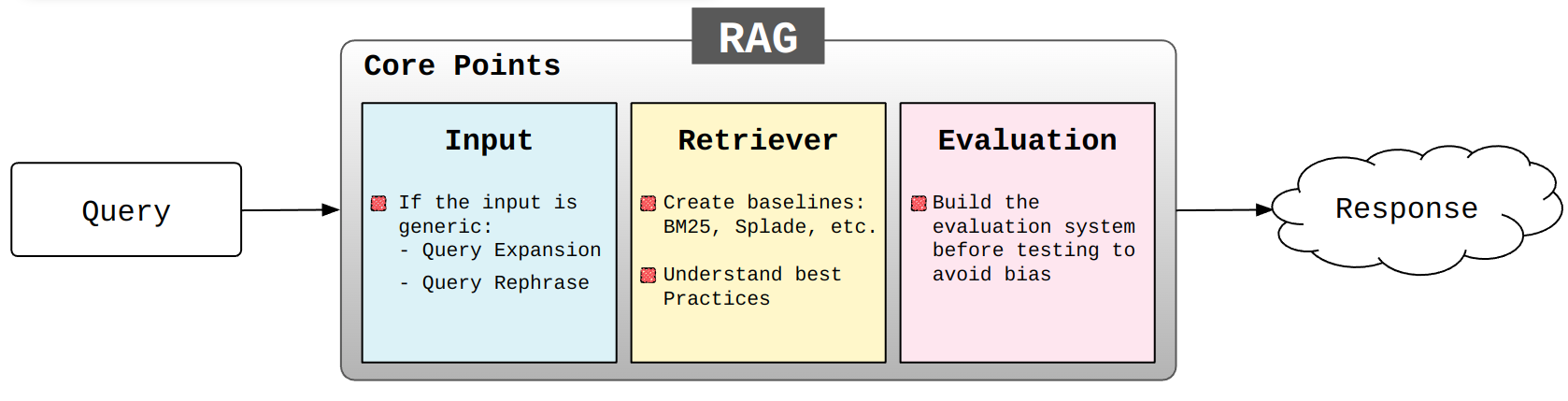} 
\caption{Core Points in RAG.}
\label{fig:rag_core_points}
\end{figure}

Despite this work being grounded in a single dataset, it is always crucial to emphasize the importance of data quality. In a simplified manner, as illustrated in Figure \ref{fig:rag_core_points}, data quality in RAG can be divided into Input, Retriever, and Evaluation.

\begin{itemize}
    \item \textbf{Input:} How are the queries formulated? Are they synthetic or generic? What is the application's purpose? Building a RAG for a chatbot differs significantly from constructing a RAG for extracting information from a long and complex document.
    \item \textbf{Retriever:} How do the data behave in information retrieval? Are the queries strongly linked by keywords in the text? What is the cosine similarity of query-to-query, query-to-document, and document-to-document?
    \item \textbf{Evaluation:} How are the data measured? Defining metrics and success rates, as in the first experiment, is always a safe path to avoid biases. Build the evaluation system before testing, avoiding bias.
\end{itemize}

% Ao final, a principal contribuição deste trabalho foi identificar e apresentar a melhor configuração possível de técnicas e parâmetros para uma aplicação de RAG. Na Figura \ref{fig:rag_openai}, são apresentados de forma resumida os resultados dos experimentos end-to-end das abordagens discutidas, e as melhores práticas recomendadas por este trabalho atingem uma acurácia final de $98.61\%$, o que representa uma melhoria de $40.73$ pp em relação ao baseline.

\begin{figure}[htb!]
\centering
\includegraphics[width=0.99\linewidth]{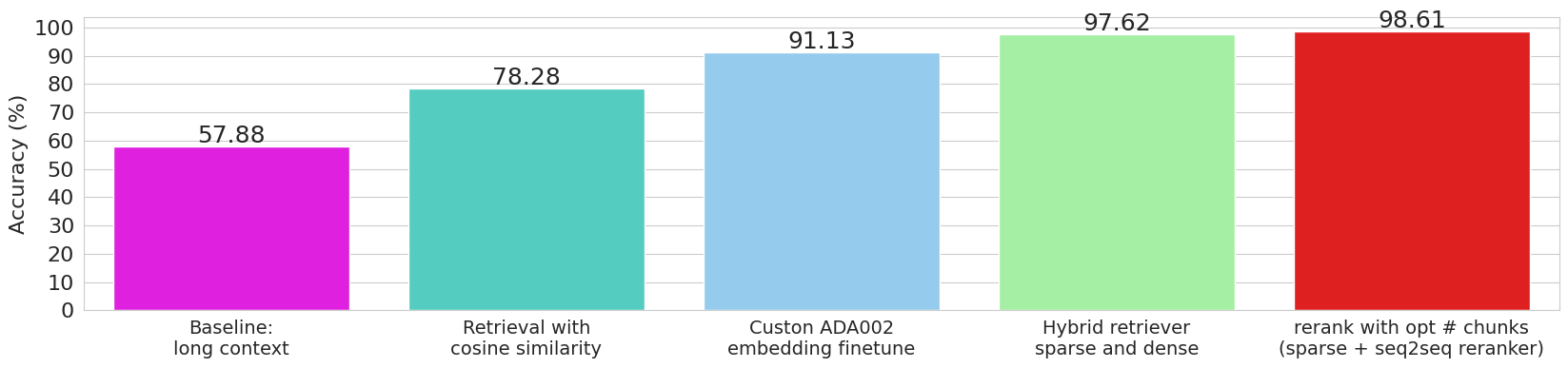} 
\caption{RAG Performance: Performance evolution on the Harry Potter dataset, where an accuracy of 100\% is considered the relative maximum.}
\label{fig:rag_openai}
\end{figure}

In conclusion, the main contribution of this work was to identify and present the best possible configuration of techniques and parameters for a RAG application. Figure \ref{fig:rag_openai} provides a summarized overview of the end-to-end experiment results for the discussed approaches, and the best practices recommended by this study achieve a final accuracy of $98.61\%$, representing an improvement of $40.73$ percentage points compared to the baseline.

\subsection{Future Work}
% In this context, our future work will be related to expanding the experiments presented here applied to datasets with real data.

We will expand our search to cover additional data sets, preferably those containing real data that already have questions and the reference documents for the answers. By using such datasets, we intend to explore techniques related to segmentation and chunk construction, as elaborated in Appendix B, Section \ref{appendix:sentence_window}.

\vspace{1cm}

\bibliographystyle{IEEEtran}
\bibliography{rag}

\newpage

\appendix
\section{PROMPTS}
\label{appendix:prompts}
All prompts were translated into English.

\begin{mybox}[colback=white]{Prompt: Make QA dataset}
\begin{lstlisting}
You will receive a document that needs you to create 1 question. It is necessary that you create only `1 question` for the document.
`I will use the created question to be answered given the entire Harry Potter book, so the question needs to be clear and always detailed to have only one unique answer throughout the book.`
## Follow exactly the following instructions, take the time necessary to produce a quality question:
1. You need to create a question similar to the `Squad` dataset, where the answer to this dataset is a span of the text. The question must be unique, and the answer must be contained within the document.
2. The answer to the question should be a faithful copy of the context of the document.
3. Create a detailed question where the answer must have more than 4 words.
4. Avoid generic questions.
The output should be a json with the characteristics of the example below:
Take your time, make sure the json does not have errors. Check and make sure there are no errors when loading this json into a Python code using `json.loads`.
{"question": "Write the question here.",
"answer": "Write the answer here. It must be a span and needs to be contained in the context."},
# Begin of the document:
{document}
\end{lstlisting}
\end{mybox}

\begin{mybox}[colback=white]{Prompt: Evaluator}
\begin{lstlisting}
You will receive two texts that need to be compared according to the scoring system described below:

# Scoring system description.
- Score 1: The response has no relation to the reference.
- Score 3: The response has little relevance but is not aligned with the reference.
- Score 5: The response has moderate relevance but contains inaccuracies.
- Score 7: The response is aligned with the reference but has minor omissions.
- Score 10: The response is completely accurate and aligns perfectly with the reference.

## Start of text 1:
{text_1}
###
## Start of text 2:
{text_2}

Evaluate with great care and thoroughly, take the time necessary to provide a quality score.

# Expected scores
- Score 1: ["only the crazy know.", "everything is divine and wonderful."],
- Score 3: ["A car has 4 wheels.", "Is there any blue car with 5 wheels."],
- Score 5: ["Harry Potter is friends with Hermione and Snape.", "Hermione has several friends, including 3 men."],
- Score 7: ["Harry Potter is friends with Hermione.", "Hermione is friends with some wizards, including one who wears glasses."],
- Score 10: ["Brazil won 5 FIFA World Cup in soccer", "Brazil is a five-time world champion."]

`Respond with only a numerical score.`
\end{lstlisting}
\end{mybox}

\section{Sentence Window}
\label{appendix:sentence_window}
There are a wide variety of techniques available for dealing with RAG and new approaches are constantly emerging. In this work we highlight the main existing strategies, although we have also explored others, such as the RAG Sentence Window approach. 

\begin{figure}[htb!]
\centering
\includegraphics[width=0.8\linewidth]{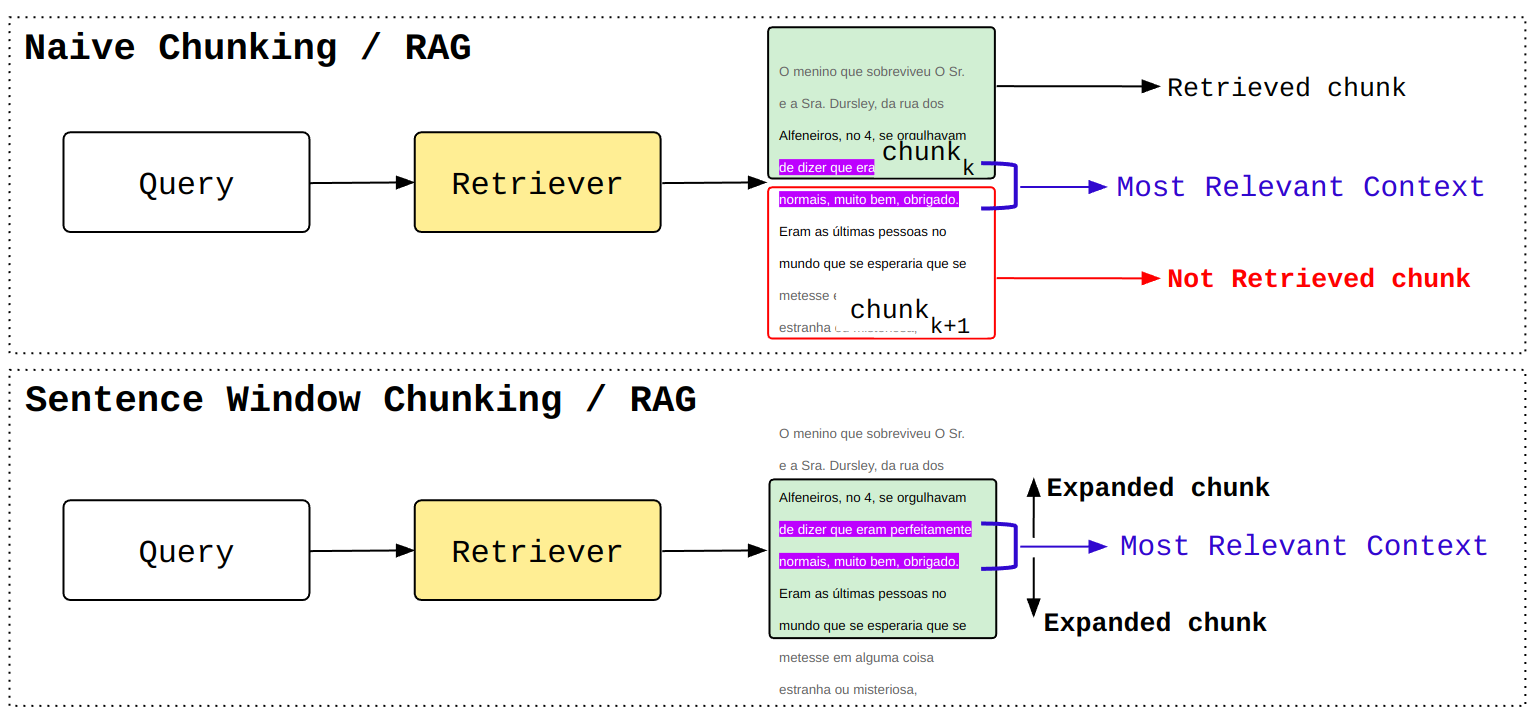} 
\caption{Comparison between the naive chunking and small to big strategies, where the latter initially retrieves a single sentence and then performs windowing with padding on both sides.}
\label{fig:rag_window}
\end{figure}

The sentence window chunk constitutes a strategy embedded in the llama-index\footnote{\url{https://docs.llamaindex.ai/en/stable/examples/node_postprocessor/MetadataReplacementDemo.html}}, which employs the small-to-large approach. The entire chunk retrieval process is based on cosine similarity. Initially, a single sentence is retrieved, followed by expanding (window) around the retrieved sentence to construct a more extensive chunk. This method implies that the most relevant context is always centered and retrieved first, as opposed to being at the edges of a chunk boundary. Figure \ref{fig:rag_window} illustrates the comparison between naive chunks and those obtained through the "sentence window" strategy.

Despite being a promising technique, the results on our dataset were inferior to the naive strategy, as indicated in Table \ref{tab:rag_window}.

\begin{table}[htbp]
\caption{Performance of the RAG Sentence Window.}
\centering\centering\resizebox{0.5\textwidth}{!}{
\begin{tabular}{lcc}
\hline
\textbf{Model}     & \textbf{Average Score} & \textbf{Degradation} \\ \hline
gpt-4              & 5.77                   & -23.5\%\\
gpt-4-1106-preview & 5.51                   & -24.7\%\\
gpt-3.5-turbo-1106 & 5.60                   & -23.8\%\\
\hline
\end{tabular}
}
\label{tab:rag_window}
\end{table}

Our hypothesis for the low performance includes:
\begin{enumerate}
    \item The construction of the question and answer occurred after the "chunk" cut, so there is no answer between two chunks. 
    \item Embedding sentence by sentence in the Harry Potter book causes ambiguity: there are multiple sentences with the name \textit{Harry Potter}, which adds a significant challenge to the step of retrieving the correct sentence.
\end{enumerate}

\end{document}